# Swarm Robots in Mechanized Agricultural Operations: Roadmap for Research

**Daniel Albiero, D. Sc.\*. Angel Pontin Garcia, D. Sc.\*. Claudio Kiyoshi Umezu, D. Sc.\*. Rodrigo Leme de Paulo, Eng.\***

*\*School of Agricultural Engineering, University of Campinas, Campinas-Brazil, ZIPCODE: 13083-875*
*Brazil (Tel: 55-19-3521-1024; e-mail: daniel.albiero@gmail.com, angelpg@unicamp.br, umezu@unicamp.br, rodrigolemepaulo@gmail.com).*

Abstract: Agricultural mechanization is an area of knowledge that has evolved a lot over the past century, its main actors being agricultural tractors that, in 100 years, have increased their powers by 3,300%. This evolution has resulted in an exponential increase in the field capacity of such machines. However, it has also generated negative results such as excessive consumption of fossil fuel, excessive weight on the soil, very high operating costs, and millionaire acquisition value. This paper aims to present an anti-paradigmatic alternative in this area. It is proposing a swarm of small electric robotic tractors that together have the same field capacity as a large tractor with an internal combustion engine. A comparison of costs and field capacity between a 270 kW tractor and a swarm of ten swarm tractors of 24 kW each was carried out. The result demonstrated a wide advantage for the small robot team. It was also proposed the preliminary design of an electric swarm robot tractor. Finally, research challenges were suggested to operationalize such a proposal, calling on the Brazilian Robotics Research Community to elaborate a roadmap for research in the area of swarm robot for mechanized agricultural operations.

Resumo: A mecanização agrícola é uma área de conhecimento que evoluiu no decorrer do último século, seus principais atores são os tratores agrícolas que, em um período de 100 anos, aumentaram suas potências em até 3.300%. Esta evolução se traduziu em um aumento exponencial da capacidade operacional de tais máquinas, mas também gerou resultados negativos, tais como: consumo excessivo de combustível fóssil, peso exagerado sobre o solo, custos operacionais muito altos e valor de aquisição milionário. Este paper pretende apresentar uma alternativa anti-paradigmática nesta área, propondo um enxame de pequenos tratores robóticos elétricos que juntos tenham a mesma capacidade operacional que um grande trator agrícola convencional. Foi realizado um comparativo de custos e de capacidade operacional entre um trator de 270 kW e um enxame de dez tratores robóticos de 24 kW cada, o resultado demonstrou ampla vantagem para o time de pequenos robôs. Também foi proposto o ante-projeto de um trator robô elétrico e finalmente foram sugeridos desafios de pesquisa para operacionalizar tal proposta, conclamando a comunidade pesquisadora em robótica brasileira em elaborar um roadmap para pesquisa na área de swarm robot para operações agrícolas mecanizadas.

*Palavras-chaves*: Capacidade de campo; Custos; Trator elétrico; Inteligência Artificial; Aração.

## 1. INTRODUCTION

Agricultural mechanization is the area of knowledge in agribusiness wich has the highest energy expenditure and the highest aggregate cost in agricultural production, reaching 60% of energy consumption according to (D Albiero, 2011). This fact occurs due to the specificities of farming operations that requires a lot of energy in the mechanical form (Goering & Hanson, 2004) referring to the different phases of agricultural production: soil preparation, seeding, planting, crop management, harvesting and conditioning of crop residues.

This energy is from power sources known as agricultural tractors (Goering et al., 2003), which enables the operation of plows, harrows, seeders, harvesters, sprayers, brush cutters, chisels, subsoilers, crushers, conditioners, rakes, terriers, planters, cutters, etc. (Srivastava et al., 2006). Since the appearance of the agricultural tractor at the end of the 19th century and the beginning of the 20st century, the power and weight of these machines have tended to increase, due to the need of improving their field capacity in the area (Goering & Hanson, 2004). For comparison, at the beginning of the 20th century, the largest tractors had approximately 15 kW of power (Renius, 2020). Today, at the beginning of the 21st century, we have reached a point of scale between the power and field capacity of agricultural tractors in the 500 kW range (Deere, 2020). There is a consensus in contemporary literature that the power growth curve of these machines is stabilizing and reaching an asymptotic limit. This trend is approaching a technological limit for parameters that represent three crucial problems. The first is the excessive energy consumption of large tractors that consume a lot of fossil diesel fuel (up to 150 liters per hour) (ASABE, 2013); the second refers to the weight of these machines, which increased from about 1300 kg in 1902 to 25000 kg in 2019 (Renius, 2020). This weight increase is necessary for the traction generated by the machine to be used, since a light tractor with great power would skate, sliding on the agricultural soil, which represents loss of efficiency (Macmillan, 2002).

Today there are tractors weighing more than 25 tons and this fact generates a very significant degradation of the soil in physical-mechanical terms which is translated into soil compaction, a phenomenon that reduces the infiltration of water in the soil, increases the force necessary for seedlings to emerge and it does not allow plant roots to go deeper (Kiehl, 1979), all of these facts represent losses in food production; and the third not least is the investment cost of these machines, which in this category (500 kW), reach values of US $ 550,000.00 (TractorHouse, 2020)

In this context, an exciting hypothesis is to change the current paradigm in agricultural mechanization to increase the field capacity of tractors by increasing their power and weight. This paper proposes a roadmap for research in the opposite direction: to decrease the power and weight of the tractors and to increase their number, optimizing the agricultural operations in terms of logistics, operational geometry, and energy efficiency: instead of using a gigantic machine of 500 kW, use 20 small tractor 25 kW.

However the problem with this anti-paradigmatic approach. He comes up against the current socio-economic situation of agricultural fields in western nations (Brazil among them) (D. Albiero et al., 2019; D. Albiero et al., 2015; D. Albiero, 2019): Tractor operators are scarce, and their costs (wages, charges, taxes, training, and insurance) are relatively high. Thus, a very suitable solution offered by the science of robotics is the operationalization of these small tractors as multiple robots operating in a swarm configuration. To economically justify such a solution and present the literature concerning swarm robots for agriculture.

The main objective of this paper is to propose that researchers of the Brazilian Robotics Research Community develop a roadmap for future research aiming to operate the use in the agricultural field of swarm robots be developed in a commercial, concrete and practical way focused on mechanized agricultural operations.

The contribution of this paper is to foster discussion about the commercial implementation of swarm robots for mechanized agricultural operations, generating discussions and perhaps initiating fruitful interactions and transdisciplinary partnerships between the robotic research community and researchers in the field of agricultural mechanization.

## 2. RELATED WORK

Robots are not new in agriculture; there is much research being developed, some of them very advanced, and already with real applications in the field, agricultural robotics is an overwhelming trend (D. Albiero, 2019). Wolfert et al. (2017) describe these advances in Agriculture 4.0, which in farming is called Smart Farming. They explain that smart machines and crop sensors on farms have obtained large amounts of agricultural data and that the quantity, quality, and scope have grown enormously, which makes data available to improve processes. In this context, innovations in the field are developing at an accelerated rate. (Bechar & Vigneault, 2016).

There are robots for the application of phytosanitary products; for sowing; for diagnosis of soil, plants, water; with computer vision systems; for harvest; with remote steering control systems; with transplant systems; for weed control; for monitoring diseases and pests; for pruning (Bechar & Vigneault, 2017).

An exciting innovation in the Smart Farms concept was a robot for irrigating pots in agricultural greenhouses. It uses sensors for humidity, position, and computer vision to assess how much each plant, individually, needs water and then performs the necessary water slide for each plant. This system makes it possible to save water and substantially improve irrigation efficiency (Araújo Batista et al., 2017). Xaud et al. (2018) developed an interesting robot for use in bioenergetic crops, De Lemos et al. (2018) present a uni-sensor strategy for navigation between rows of crops for robots and Oliveira et al. (2018) presented a methodology to adapt conventional commercial systems to autonomous robotic systems.

Davis (2012) described a family of agricultural vehicles that has collective sensing and computational infrastructure. There is an exciting European research program that deeply studies applications of swarm robotics concepts with UAVs used to obtain information on the productivity of beet fields and to generate data on weeds, diseases, and nematodes (Toorn, 2020). Albani et al. (2019) use UAVs swarm robots to monitor and map weeds in agricultural fields. Albani et al. (2017) presented an exciting roadmap for future studies on swarm robotics for applications in farm monitoring and mapping. Mukherjee et al. (2020) studied the challenges in operationalizing the use of UAVs in swarm robotics configuration to operate decentrally and heterogeneously in the agricultural environment, which has very variable and non-trivial control edges.

Shamshiri et al. (2018) have made an extensive literature review on agricultural robotics; its challenges and special attention is given to multi-robots and swarm methods. (Blender et al., 2016) introduced Mobile Agricultural Robot Swarms (MARS) is an approach for autonomous farming operations by a coordinated group of robots and describes an application in seeding. Trianni et al. (2016) described the concept of a set of swarm robots for weed control and define a roadmap for the execution of such a project. Minßen et al. (2017) presented conceptual studies for agricultural care in plants considering several swarm robots.

The aforementioned papers present the current state of the art on the theme of swarm robots for agricultural applications, when analyzing the issues of problems studied and solved by the authors it is noticed that there is still much to be accomplished, many important solutions for the commercial operationalization of a swarm of robots operating in the field has by no means been resolved, in this context there is a huge opportunity for developments, which is very motivating.

## 3. OPERATIONAL AND COST COMPARATIVE

To carry out the operational and cost comparison between the configuration of small power swarm robots and a large tractor, one of the most power demanding agricultural operations was chosen: deep plowing in loose clayey soil with a moldboard plow. According to the ASABE D497 standard (ASABE, 2013), the request for the tractive force for such an operation is given by (1):

$$D = F_i . [A + B.(S) + C.(S)^2]. W.T \quad (1)$$

Where: D is the tensile strength of the implement (N);

A, B, C are dimensionless coefficients tabulated by the D497 standard;

S is the travel speed (km s$^{-1}$);

W is the cutting width of the implement (m);

T is the cutting depth of the implement (cm).

In this paper, a plow of 5 molds by Marchesan, model ARR2, was chosen for the large tractor, in double composition by a tandem header, making ten molds, cutting width of 4 m and cutting depth of 0.35 m. For the small robot, the same plow with one moldboard was chosen, which configures a cutting width of 0.40 m and depth of cut of 0.35 m.

According to the D497 standard, the typical displacement speed of this plow is 5 km/s, the dimensionless coefficients are A = 652; B = 0; C = 5.1. For soil with a clay texture, the Fi factor is 1. The following data are obtained:

$D_{10aivecas}$=109,130.00 N; $D_{1aiveca}$=10,913.00 N.

Considering loose soil, the large tractor with the power source from the internal combustion diesel engine coupled to a mechanical transmission system in the MFWD system, according to the D497 standard, has an overall efficiency in the transfer of tractor/plow power of 53.9%. In this work, agricultural swarm robots will be considered as a result of research by Melo et al. (2019) and Vogt (2018) that designed and dimensioned a small electric tractor powered by electrochemical power from batteries and electric engines direct drive in tracks on a rubber mat. In this configuration, the overall efficiency in the transfer of tractor/plow power is 76.4%.

The equation (2) gives the nominal power required in tractor engines (conventional for the large tractor and electric tractor for the small swarm robot tractor):

$$P_{nom} = \frac{D.S}{\eta} \quad (2)$$

Where: D is the tensile strength (N);

S is the travel speed (m s$^{-1}$)

η is the overall tractor/implement efficiency (decimal).

So, we have the following data:

$P_{nom}$ (Large Tractor) = 281.20 kW;

$P_{nom}$ (Swarm Tractor) =19.83 kW;

The John Deere 8730R large tractor was selected ((Deere, 2020). With a nominal power of 272 kW and a maximum power of 300 kW. Maximum weight with ballast of 19.805 kg. For the Swarm Electric Robot Tractor (TRSE) recommendations of (Melo et al., 2019; Vogt, 2018; Vogt et al., 2018), the drivetrain being sized with two 10 kW electric motors each coupled to the wheels, the power source comes from a quick replacement pack consisting of 4 stationary lead-acid batteries 12V/220 Ah per battery, 2.5 hours autonomy for each pack and total weight of the 700 kg electric swarm robot tractor.

Considering the operational efficiency of the tractor/plow set of 70% (ASABE, 2013), the field capacity of the set can be calculated by (3):

$$Cc = \frac{S.W.ef}{10} \quad (3)$$

Where: Cc is the field capacity (ha h$^{-1}$);

S is the travel speed (km h$^{-1}$);

W is the cutting width (m);

ef is the operational efficiency (decimal).

We have the following data:

Cc (Large Tractor) = 1.4 ha h$^{-1}$;

Cc (Swarm Tractor) = 0.14 ha h$^{-1}$

In this work, the operating cost for the large tractor will be considered only the composition between the cost of diesel fuel and the cost of the operator (salary, charges, insurance, and training), maintenance, depreciation, financial, and investment costs will not be considered.

The fuel consumption cost can be obtained according to the D497 standard considering the diesel consumption obtained by (4):

$$Cf = 2.64\,X + 3.91 - 0.203\,\sqrt{738X + 173} \quad (4)$$

Where: Cf is the diesel consumption (L kW h$^{-1}$);

X is the ratio between the power in the Power Take Off (PTO) equivalent for the agricultural operation and the total power of the PTO (decimal).

In deep plowing operation, the ratio X is equal to 1. Therefore, the estimated fuel consumption of the John Deere 8730R tractor, according to the D497 standard (ASABE, 2013) is:

Cf = 0.42 L kW h$^{-1}$

Considering a working period of 1 hour at the tractor's nominal power, there is a consumption of 114.24 liters of diesel. With an exchange value for the dollar in January 2020 of R$ 4.16 per US $, the amount of one liter of diesel in January 2020 was US $ 0.77. Therefore, the hourly fuel cost for the large tractor is US $ 87.96. The same dollar quote was used to convert all costs.

About the operator's monthly cost, in a 40-hour workday, according to data from (BRASIL, 2020; CNA, 2020), the median salary of a tractor operator in Brazil is US $ 361.53 in the exchange rate. January 2020. In addition to this amount, there are labor charges, approximately 70% of the salary (Fernandes, 2020) US $ 252, the cost of life insurance US $ 13.90 (CNA, 2020) and costs with training US $ 6.27

(SENAR, 2020). In total, there is an operator cost value of US $ 634.40 per month. Per hour the value is US $ 15.86.

Therefore, the operating cost of the large tractor is US $ 103.82 per hour.

Regarding TRSE, there is no cost for the operator. Only the one related to the electrical system and the charging of energy from the grid. This cost is around US $ 2.72 per hour (Vogt, 2018). It doubles if the second pack of 4 batteries is considered to increase the system's autonomy to 5 hours.

A 272 kW tractor with operating capacity ten times greater than a small 20 kW TRSE has a much higher cost. However, when considering a set of 10 multi-robots operating according to swarm methods, the field capacity is equalized. In this case, the total cost of the ten multi-robots would be around US $ 27.20 per hour, considering 5 hours of autonomy. With the second battery pack, it's cost is US $ 54.40 per hour, half the hourly cost of a large tractor.

The purchase price of a John Deere 8730R tractor (TractorHouse, 2020) is US$ 355,400.00. For cost estimation this paper proposes a multifunctional agricultural swarm robot (TRSE) through the integration of robotic technologies with a new version of the electric tractor developed by (Vogt, 2018). The TRSE value composition is shown in Table 1. Figure 1 and 2.

**Table 1. TRSE value composition.**

| Component | UN. | Unit value (US$) | Value (US$) | Ref. |
|---|---|---|---|---|
| Electric Motor HPEVS AC23-96V/650A | 2 | 3,800.00 | 7,600.00 | (HPEVS, 2020) |
| Controller electric motors Curtis1238-96V | 1 | 2,150.00 | 2,150.00 | (Curtis, 2020) |
| PAC controller for tract or access/power systems Advantech APAX5620KW | 1 | 1,130.00 | 1.130,00 | (Advantech, 2020b) |
| Processor Artificial Intelligence Intel Core-i7 processor | 1 | 673.00 | 673.00 | (Terabyte, 2020) |
| Sensoring | - | - | 2,000.00 | - |
| Communication to power systems module Advantech APAX5490 CANbus communication | 4 | 172.00 | 688.00 | (Advantech, 2020a) |
| Communication CAN module Curtis 1351 for Trimble EZ and Controller Curtis 1238 | 1 | 198.00 | 198.00 | (*Motor Controllers | Curtis Instruments*, n.d.) |
| Automatic Pilot GNSS Trimble EZ | 1 | 7,250.00 | 7,250.00 | (Trimble, 2020) |
| Tractor Chassis/Tires | - | - | 800.00 | - |
| Hydraulic systems | - | - | 600.00 | - |
| Mechanical Drivetrain/tracks | - | - | 500.00 | - |
| Power systems (PTO, hydraulic arm, drawbar) | - | - | 500.00 | - |
| Battery Moura 12MS234 12V/220Ah | 4 | 352.00 | 1,425.00 | (WinnerShop, 2020) |
| | | **Total** | **25,316.00** | |

Summarizing this comparison, we have that a 10 TRSE swarm has the same field capacity in hectares per hour in plowing as the John Deere 8730R tractor (1.4 ha/h). However, the cost of purchasing a JD8730R is US $ 102,240.00, more than the sum of the value of 10 TRSE. In the comparison of operating cost, for the same field capacity, the cost of the JD8730R is 3.2 times higher than the operating cost per hour of the TRSEs swarm. And the weight of the large tractor is 2.8 times greater than the total weight of the 10 TRSE, but a caveat is necessary. In terms of soil mechanics, the value to be considered in the operation of the TRSE is 7000 N, as it's the request that the soil receives in compression from an electric robot swarm tractor. In the region where a TRSE passes, no other will pass, since the operation was carried out. There is no need for traffic on the ground. On the other hand, in the region where a JD8730R passes, the soil undergoes a compression equivalent to 198.050 N. The effects soil compaction will be completely different, and favorable to TRSE.

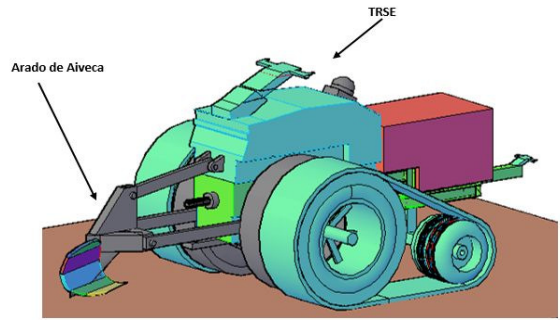

Figure 1. Tractor Robot Swarm Electric (TRSE) simulating the use of a moldboard plow.

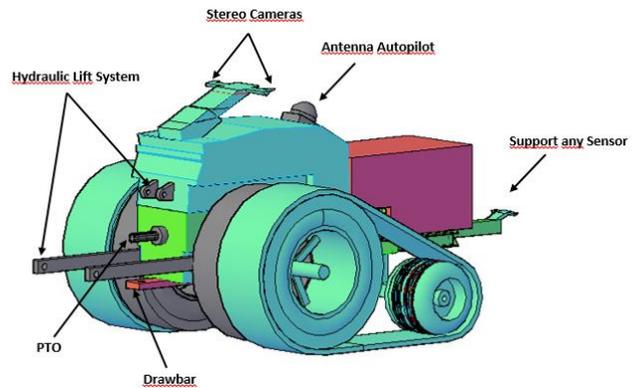

Fig. 2. Tractor/Implements Power Transfer Systems

A critical issue in this context is the impact of the widespread use of robotic systems in the agricultural field in terms of personnel with technical knowledge to operationalize this type of vehicle and perform the maintenance of equipment. This challenge may be as difficult as developing robot swarms. However, the authors believe that if Brazil wants to continue to be a major player in global agribusiness, it has no alternative but to invest in education and training for its workforce (notably operators conventional tractors) is transformed according to the new world trend of agriculture 4.0, which has one of its fulcrums in robotization.

Institutions like National Rural Learning Service (SENAR, 2020) must pay attention to this need and train frontline workers, Federal Institutes (*Instituto Federal*, 2020) need to train the necessary technicians and universities to develop research and train engineers able to implement them on the farms.

## 4. CHALLENGE IN AGRICULTURAL SWARM ROBOT: A ROADMAP FOR RESEARCH

This draft of roadmap can generate very fruitful interinstitutional and transdisciplinary partnerships that will be able to implement innovative and essential research so that this proposal goes off the record and becomes one carried out in the agricultural world. In this context, it is important to emphasize that there is a huge field for studies with great challenges concerning the operationalization of each of the agricultural operations (Minßen et al., 2017) within the universe of robotics, specifically in the area of multi-robots working in swarm methods. The ASABE D497 standard defines 48 different types of agricultural operations with different characteristics and parameters (ASABE, 2013). For each of them, several challenges for the operationalization of a swarm robot system are presented and are suggested below:

*4.1 Behavior –base systems*

The great challenge of this line of research is to develop control architectures based on behavior that can be adapted to the unstructured agricultural environment, new methods of incremental learning must be developed and adapted of robots based on war environments or catastrophe environments. In this context, reinforcement learning is an excellent methodology to optimize agricultural robotic systems based on behavior, because through the "decomposition" of the behavior in small sub-behaviors it is possible to reduce the size of the phase space effectively. This is another major challenge in this line research to find the best learning network through which the modularization of learning "policies" results in accelerated and more robust learning.

*4.2 AI reasoning methods*

The essential question for robotics in the elaboration of the application of artificial intelligence methods is to define which are the suitable formats for KR, and from this definition find the state function that refers to generation and maintenance, in real-time, of a symbolic description of the robot's environment, based on a recent situational condition of the environmental information obtained by sensors and communication with other agents involved in such a way that decision-making is correct and optimized for solving a problem or overcoming a barrier.

*4.3 Collective-level behavior*

The behavior at the collective level of the swarm is what defines the conclusion of the global mission about each specific objective of the agricultural operation, the challenge here is to develop swarm robotics techniques that enable the emergence of emergent group behaviors, such as self-organization, flexibility in joint operations, and scalability in terms of common objectives.

*4.4 Operational strategy*

The division of the agricultural field can be configured in cells, lines, bands, blocks, in short, several sets with different topologies, which in the real agricultural environment take very complex forms due to the specificities of the relief, contour lines, shape of the fields, planting configuration of cultures. All these topological parameters lead to complex logistical problems for the optimization of the movement strategy of the elements of the swarm, reaching questions related to differential geometry.

*4.5 Stochastic process computing*

A significant challenge for the operationalization of a robotic tractor swarm in the field is the extreme unpredictability that the agricultural environment has. Even in a homogeneous culture sown with a uniform pattern, it has a high variability of shapes, geometries, positions, and scales. This fact occurs due to the treatment of living elements, which interact with the climate, soil and other living beings (micro and macroscopically), in this concept it is necessary to enter into the area of stochastic processes so that there is a better understanding of the operational strategies as well as an adaptation of the decision-making algorithms against random components.

*4.6 Multiple decision-making*

In agricultural swarm robotics its necessary the definition of the algorithms that can be used, because of the processing capacity of the machines about the extraordinarily complex and multiple decision-making problems required in unstructured agricultural environments and which has objects very fragile and variable (living beings).

*4.7 On-board systems*

Computer vision systems are a universe, from the development of specific hardware to the elaboration of suitable firmware. When thinking about the immense range of sensors, receivers, and transducers necessary to make a swarm robot operational in the field, it is essential to divide this field of studies into the proprioception, exteroception, and guidance system. The challenges are the sensors optimization of the sensors and actuators, necessary to enable the internal and external robot's operation, depending on the specific agricultural process. In terms of guidance, the agricultural environment offers immense obstacles, from varying light conditions, such as irregular reliefs, to complex and often discontinuous contour geometry.

*4.8 Hardware enhancement*

According to (D. Albiero, 2019) the main obstacle is the development of adapted to the agricultural conditions systems, because there are many very good elements of automation and robotics used in industry and smart cities, but when they are part of the agricultural world (with high susceptibility of the agricultural products in the spoiling of the most varied forms), problems occur. There is the urgent necessity of developments in robotics technologies for agricultural reality.

*4.9 Networked Swarm robots*

This line of research is very challenging because, through the connection between the members of the swarm, it is possible, through distributive computing, to imitate the behavior of decentralized animals through simple behaviors, capable of generating complex responses at the collective level, the so-called emergent behaviors.

*4.10 Distributive computing*

In particular, the concept of parallelism has become essential; we have seen the development of multicore CPUs. With a swarm of robots in the agricultural field, the distributed computing system functionality is immense, both in terms of capturing data and generating useful information to optimize the specific function performed. The challenge is to integrate all this processing through wireless communication networks in the field.

*4.11 Particle Swarm optimization*

According to (Nedjah & Macedo Mourelle, 2006), particle swarm optimization is a mathematical optimization method that mimics the behavior of insect swarm, the challenge is to find the optimal path or solution for the entire swarm and implement communication between the swarm robots network optimizing the answer ahead of the swarm before the common goal. This line of research has deep interfaces with lines 5.1, 5.3, 5.9, and 5.10.

## 5. CONCLUSION

The comparison of costs and field capacity between a set of Swarm Electric Robot Tractors (TRSE) and a large tractor was performed, demonstrating the feasibility of the swarm configuration to replace the large tractor for deep plowing. The constructive preliminary design of a new autonomous 24 kW electric robot tractor within the swarm operating methodologies have been described. Challenges for future research focused on the implementation in the agricultural field of swarm robots aiming at mechanized operations are suggested. This initiative can constitute a roadmap for interinstitutional and transdisciplinary research bringing together the scientific community dedicated to robotics in Brazil.

## ACKNOWLEDGMENT

The authors thanks the productivity grant of CNPq and the physical and financial resources made available by FUNCAP, CAPES, CNPq, UFC, and UNICAMP.